\documentclass[final]{adacs}
\usepackage{times}
\usepackage{epsfig}
\usepackage{graphicx}
\usepackage{amsmath}
\usepackage{amssymb}
\usepackage[numbers]{natbib}
\usepackage{import}

\def\red{\textcolor{red}}
\def\blue{\textcolor{blue}}
\usepackage{multirow}
\usepackage{threeparttable}
\usepackage{algorithm, algorithmic}

\usepackage[pagebackref=true,breaklinks=true,colorlinks,bookmarks=false]{hyperref}

\renewcommand{\paragraph}[1]{{\vspace{+0mm}\noindent\bf #1}.}

\begin{document}


\title{Adaptive and Cascaded Compressive Sensing}

\vspace{-2mm}

\author{Chenxi Qiu \qquad Tao Yue \qquad Xuemei Hu \\
Nanjing University\\
{\tt\small chenxiqiu@smail.nju.edu.cn, \{yuetao, xuemeihu\}@nju.edu.cn}
}


\maketitle


\begin{abstract}
Scene-dependent adaptive compressive sensing (CS) has been a long pursuing goal which has huge potential in significantly improving the performance of CS. However, without accessing to the ground truth image, how to design the scene-dependent adaptive strategy is still an open-problem and the improvement in sampling efficiency is still quite limited.
In this paper, a restricted isometry property (RIP) condition based error clamping is proposed, which could directly predict the reconstruction error, i.e. the difference between the currently-stage reconstructed image and the ground truth image, and adaptively allocate samples to different regions at the successive sampling stage. Furthermore, we propose a cascaded feature fusion reconstruction network that could efficiently utilize the information derived from different adaptive sampling stages. The effectiveness of the proposed adaptive and cascaded CS method is demonstrated with extensive quantitative and qualitative results, compared with the state-of-the-art CS algorithms.
\end{abstract}


\section{Introduction}
\label{sec:intro}
Compressive sensing (CS) has attracted wide attention since it was first proposed in 2006~\cite{donoho2006compressed}, for it can greatly reduce the sampling rate required for accurate reconstruction. Thus, CS has been applied in many fields, including but not limited to image acquisition~\cite{duarte2008single, sankaranarayanan2012cs}, accelerated magnetic resonance imaging (MRI)~\cite{lustig2007sparse} and wireless broadcast~\cite{li2013new}, etc.
\begin{figure}[t]
\setlength{\belowcaptionskip}{-3mm}
\centering
\includegraphics[width=\linewidth]{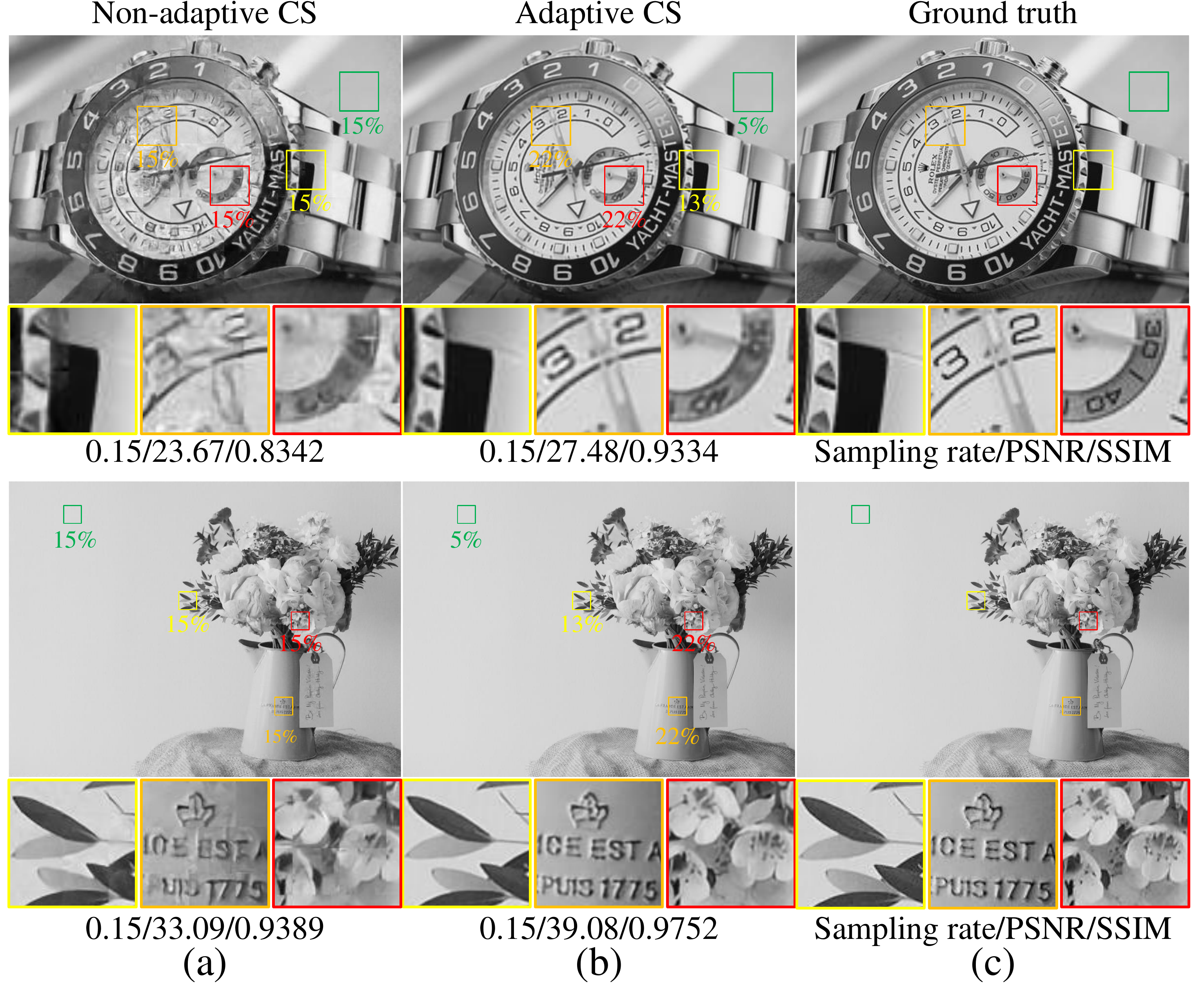}
\caption{The non-adaptive CS method ISTA-Net+~\cite{zhang2018ista} adopts the same sampling rate for each region of the target scene, which is redundant for some regions, but not enough for other regions. The proposed adaptive CS method can adaptively adjust the sampling rate of different regions according to the characteristics of the target scene, so as to improve the image reconstruction quantity. The percentage number of different colors in the figure represents the sampling rate of the corresponding framed region.}
\label{fig: fig1}
\end{figure}

Traditionally, the CS problem can be formulated as
\begin{equation}
\setlength{\abovedisplayskip}{4pt}
\setlength{\belowdisplayskip}{4pt}
\label{eq:cs_model}
\arg \min_{x\in\mathbb{R}^n}\|\Psi x\|_1\quad \textnormal{s.t. }
\Phi x + \epsilon =y,
\end{equation}
where $\Phi\in R^{m\times n}$ is the compressive measurement matrix. $\epsilon\in R^m$ and $y\in R^m$ denote the measurement noise and the compressive measurement. The number of measurements $m$ is commonly much smaller than $n$ and $\Psi$ denotes the sparse transform operator. In this model, the sampling matrix is fixed and predefined, with spatially uniform sampling rate for the entire image. However, depending on the image content, different image regions could be of different complexities, requiring quite different sampling rates for similar reconstruction quality. As shown in Fig.~\ref{fig: fig1}, the original image patch can be well recovered at a very low sampling rate for the smooth regions such as the wall and the background. However, a higher sampling rate is required for the regions with more complex textures, such as the pointer of the dial or words on the vase. Therefore, scene-dependent adaptive sampling is quite-promising in achieving much higher sampling efficiency than traditional CS methods.

Adaptive CS has been a long pursuing goal for efficient and adaptive information sampling. The most challenging part in adaptive CS is how to design the adaptive strategy when the ground truth image is not accessible. Previous methods~\cite{rousset2016adaptive, wang2019iterative, phillips2017adaptive} propose to focus on sampling regions with high feature density, according to the intermediate restored image from existing measurements. However, these methods are to-some-extent heuristic and the improvement in CS efficiency are still limited. 

In this paper, enlightened by the restricted isometry property (RIP) in CS theory, we propose an efficient adaptive and cascaded CS neural network (ACCSNet).

By utilizing the RIP-condition based clamping on the preceding measurement error maps, the unknown reconstruction errors can be revealed by the known measurement errors. Then, the error-guided adaptive scheme is proposed to realize the adaptive sampling. A cascaded reconstruction network is proposed to collaborate with the multi-stage adaptive sampling strategy. With the stage-to-stage feature fusion mechanism, information among stages can be fully fused and exploited, further improving the quality of reconstruction results. Through comparison with the state-of-the-art CS algorithms, we demonstrate the effectiveness of the proposed method both quantitatively and qualitatively. 

In all, the contributions of the proposed method are concluded as below.
\begin{itemize}
  \vspace{-2mm}
\item We propose ACCSNet to implement adaptive compressive sampling based on the RIP property in CS theory, which enables to locate and adaptively sample image regions with high reconstruction errors and realize CS imaging with high-efficiency. 
  \vspace{-2mm}
\item We propose a cascaded information fusion method in the CS reconstruction neural network which could facilitate sufficient information exploration among different stages of adaptive CS and further improve the CS reconstruction quality. 
  \vspace{-2mm}
\item The effectiveness of the proposed method is extensively demonstrated through comparing with state-of-the-art methods and ablation study. 
  \vspace{-2mm}
\end{itemize}

\section{Related Work}
\label{sec:related}
From conventional optimization methods~\cite{daubechies2004iterative, boyd2011distributed} to deep-learning based reconstruction methods~\cite{yang2016deep, zhang2018ista, metzler2017learned, kulkarni2016reconnet,xu2018lapran, yao2019dr2, chen2020learning, shi2019scalable}, the performance in CS imaging have been significantly improved. Most of these methods take the samples without considering scene contents, leaving the room for improvement in the performance.

\paragraph{Adaptive Sampling CS imaging}
Conventionally, to realize adaptive sampling, compressive measurements are sampled in different stages and the sampling region of the next stage is decided based upon the analysis of the reconstructed results of the previous stages.
To decide the sampling region at the next stage, different heuristic methods are proposed to analyze the image characteristic of current-state reconstruction and locate the image region of high density in edges, wavelet coefficients or discrete cosine transform (DCT) coefficients, etc. 
The adaptive sampling scheme based on wavelet transform~\cite{abetamann2013compressive, yu2014adaptive, soldevila2015high, rousset2016adaptive, wang2019iterative} adopts cascaded sampling and reconstruction method, which starts from the lowest resolution and obtains a higher resolution image after each stage of sampling and reconstruction. The sampling of each stage is based on the wavelet transform coefficients of the previous low resolution reconstruction results. With these adaptive strategies, the regions with large wavelet coefficients are sampled, and the regions with small wavelet transform coefficients are often discarded. 
Similar ideas are also applied to other transform domains, e.g.  DCT~\cite{liu2017adaptive}, Fourier transform~\cite{jiang2017adaptive}.
However, since regions with fine textures can often be neglected in the case of low resolution or low quality, incorporating only the feature characteristics of the intermediate reconstruction to guide the adaptive sampling process is inaccurate and prevents CS imaging of complex images with high accuracy.
Recently, inspired by the animal visual mechanism, Phillips \etal~\cite{phillips2017adaptive} proposed an adaptive foveated single-pixel imaging system, which can adaptively sample with a foveated pattern centered at the position of the moving target in the video streams. However, it is for adaptive video CS, which can not be applied to single image CS.

In all, how to design the scene-dependent adaptive CS imaging method without access to the ground truth image is still an open problem, and the potential of the scene-dependent information is still not well-explored with existing adaptive methods.
In this paper, we propose an adaptive CS method for single image CS imaging. Through fully exploiting the RIP theory, we propose an adaptive CS imaging method that could directly locate the image regions with high reconstruction error, leading to highly efficient adaptive CS imaging.

\paragraph{Cascaded Compressive Reconstruction}
Cascaded CS reconstruction methods propose to reconstruct images of different sampling rates with the same network, which greatly increase the scalability of the imaging system. Xu \etal~\cite{xu2018lapran} proposed a multi-stage Laplacian pyramid reconstructive adversarial network named LAPRAN, which can generate images of different resolutions following the concept of Laplacian pyramids.
Shi \etal~\cite{shi2019scalable} proposed an end-to-end cascaded structure for simultaneously training the sampling matrix and reconstruction network. With the scalable structure and greedy method, multiple sampling rates can be realized with only one network with high quality.
However, these multi-stage methods do not take the full advantage of the information among stages, only the sampling results~\cite{xu2018lapran} or the reconstruction results~\cite{shi2019scalable} are fused between adjacent stages, and all the feature maps generated by the different stages are discarded. In this paper, we propose a cascade information fusion method that could enforce sufficient information fusion among different adaptive stages and further improve the efficiency of adaptive CS imaging.

\section{Adaptive and Cascaded CS Net}
\label{sec:proposed}
\begin{figure*}[htbp]
\setlength{\abovecaptionskip}{-1mm}
\setlength{\belowcaptionskip}{-3mm}
  \centering
  \includegraphics[width=\linewidth]{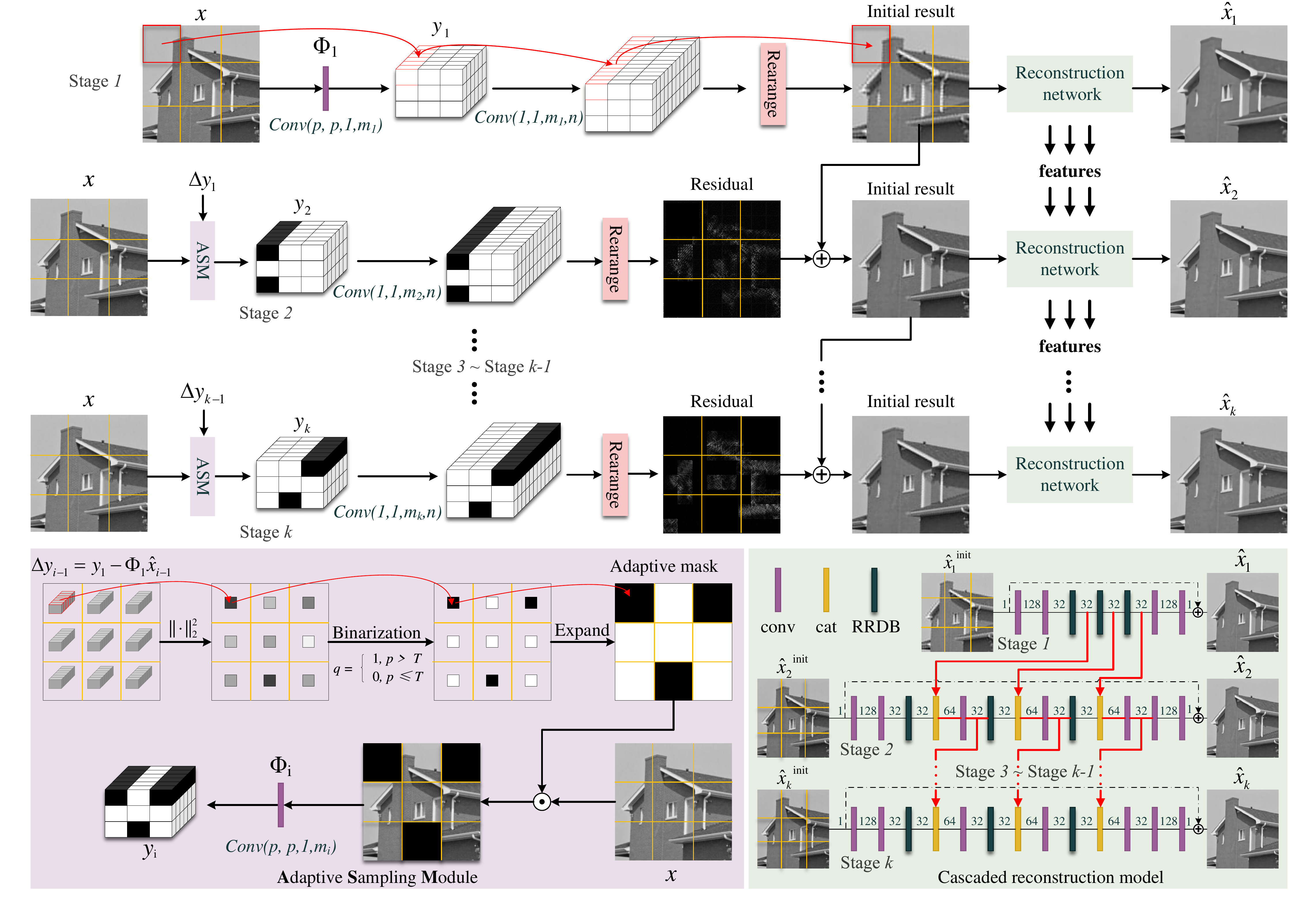}
  \caption{The diagram of \emph{k}-stage ACCSNet.}
  \label{fig: network_structure}
\end{figure*}
Different from the traditional CS model in Eq.~\ref{eq:cs_model}, we propose to model the CS problem with an multi-stage adaptive and cascaded model,
\begin{equation}
\setlength{\abovedisplayskip}{2pt}
\setlength{\belowdisplayskip}{2pt}
  \begin{aligned}
  y_{i} & = \Phi_{i} x + \epsilon\\
  \hat{x_i} & = \mathcal{N}_{i}(y_1, y_2, ..., y_i) \\
  \Phi_{i+1} & = \mathcal{H}(\Delta{y_i}), \\
  \end{aligned}
  \label{eq:accs_model}
\end{equation}
where $x$ is the latent image, $y_i$ is the measurement vector of \emph{i}-th stage, $\Phi_i$ is the corresponding sampling matrix, $\mathcal{N}_{i}(\cdot)$ indicates the reconstruction mapping function of \emph{i}-th stage and is implemented by the cascaded feature fusion network, $\mathcal{H}(\cdot)$ denotes the adaptive sampling module that is realized by the error-induced adaptive sampling module. Expanding the recursive expression in Eq.~\ref{eq:accs_model}, all modules of multi-stages combine an entire network named ACCSNet. Fig.~\ref{fig: network_structure} shows the structure of the \emph{k}-stage ACCSNet, which can realize different sampling rates and multi-stage reconstruction within only one model. 

As for the sampling matrix $\Phi_i$, we use a patch-based learnable scheme. The input image $x$ is divided into several non-overlapping $p\times p \times c$ patches $x^j$, where $p$ and $c$ denote spacial size and the number of input channels, respectively. In our paper, only grayscale images are applied for the experiment, so the input channel is set to 1. In practice, we use one learnable convolution layer \emph{conv(p, p, 1, m)} without bias to implement the sampling model, in which the kernel size is $p \times p$, the stride is also $p$, and the number of output channels is $m$. Assuming that the sampling rate of this patch is $r$, then $m$ can be calculated as $\lfloor rp^2 \rceil$. $\lfloor . \rceil$ represents the rounding calculation.

Adaptive sampling module (ASM) takes $\Delta y_{i}$ of the stage \emph{i} as input and outputs a binary adaptive mask to determine which regions should be sampled at the stage \emph{i+1}, as shown in the bottom left of Fig.~\ref{fig: network_structure}.  
Except for the first stage, the sampling model of each stage adaptively adjusts the sampling strategy according to the sampling and reconstruction results of the previous stage. 
The reconstruction model of each stage will continuously fuse feature maps of the previous stage to make full use of the information sampled before, as shown in the top and the bottom right of Fig.~\ref{fig: network_structure}. In the following, we will introduce the proposed ACCSNet in detail.

\subsection{Error-induced Adaptive Sampling Module}
\begin{figure}[htbp]
\setlength{\belowcaptionskip}{-3mm}
\centering
\includegraphics[width=\linewidth]{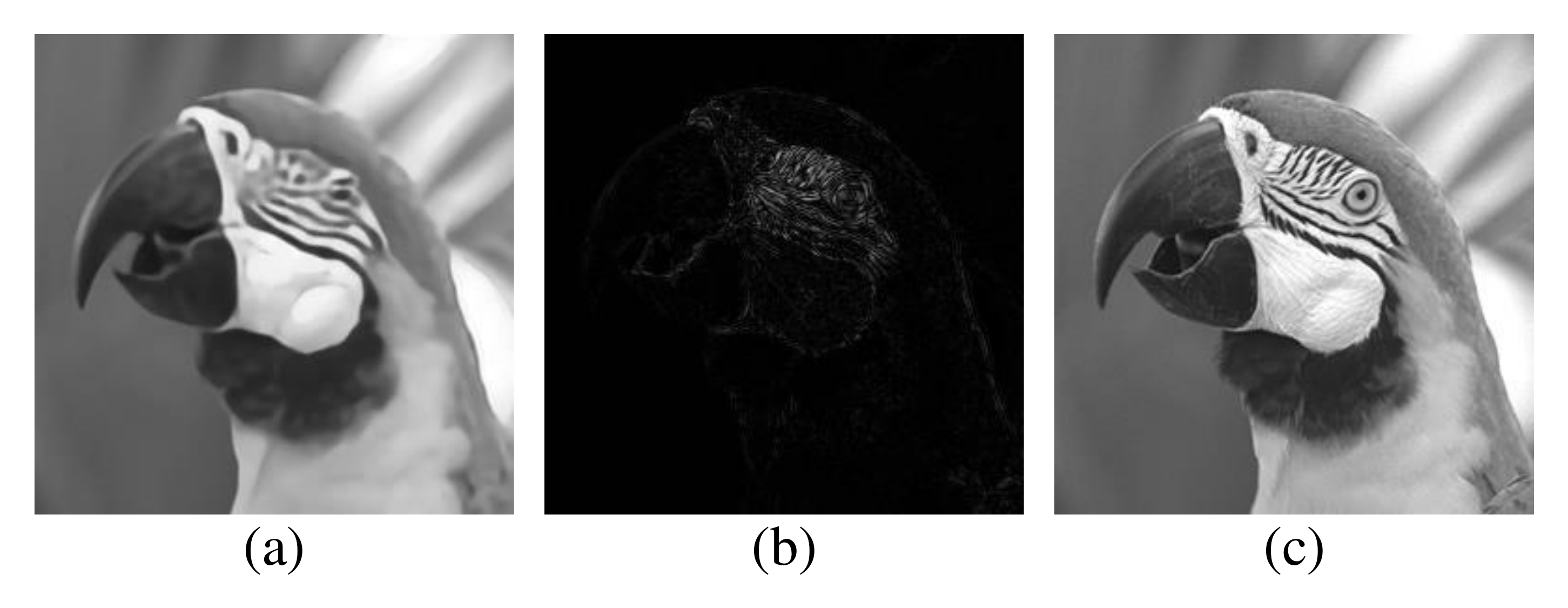}
\caption{Reconstruction results and absolute error image with sampling rate $= 0.05$. (a) reconstruction results, (b) absolute error map, (c) ground truth image in Set11~\cite{kulkarni2016reconnet}.}
	\label{fig: res_residual_gt}
\end{figure}
The core idea of adaptive sampling is how to choose the next sampling strategy according to the existing information.
Fig.~\ref{fig: res_residual_gt} shows an intuitive example, one intermediate reconstruction result of Set11~\cite{kulkarni2016reconnet} at the stage \emph{1} with sampling rate $r = 0.05$, the corresponding reconstruction error map and the ground truth image are present in Fig.~\ref{fig: res_residual_gt}(a-c) respectively. Obviously, the reconstruction error of this uniformly sampled case is not uniform. Therefore, on the next sampling stage, if we can allocate more samples to the regions with larger error, the entire sampling scheme could be more efficient. In other words, we can achieve better sampling efficiency and thus lead to higher reconstruction quality with the same total sampling rate.

\begin{figure}
\setlength{\belowcaptionskip}{-3mm}
\centering
\includegraphics[width=\linewidth]{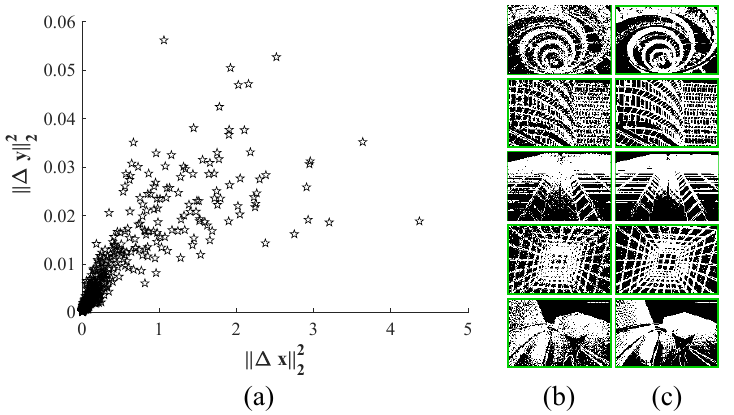}
\caption{Visual display of the relationship between $||\Delta y||_2 ^2$ and $||\Delta x||_2 ^ 2$. (a) statistical diagram of $||\Delta y||_2^ 2$ and $||\Delta x||_2^ 2$, (b) binary $||\Delta y||_2^ 2$, (c) binary $||\Delta x||_2^ 2.$}
	\label{fig: dx_dy}
\vspace{-5mm}
\end{figure}

\paragraph{RIP condition based error clamping}
The key problem of adaptive sampling is how to estimate the reconstruction error $||\Delta x_i||_2$ of the stage \emph{i} without the ground truth image?
According to the relationship between the sampling measurements and reconstruction results, we can calculate the measurements error from reconstruction error by
\begin{equation}
\setlength{\abovedisplayskip}{4pt}
\setlength{\belowdisplayskip}{4pt}
  \Delta y_{i} = \Phi_{1}\Delta x_{i} = \Phi_{1}(x - \hat{x}_{i}) = y_{1} - \Phi_{1} \hat{x}_{i},
\end{equation} 
where $y_1$ and $\Phi_1$ are sampling measurements and sampling matrix of stage \emph{1}, $\hat{x_{i}}$ denote the reconstruction results of the stage \emph{i}, and $x$ is the ground truth image, respectively.
$\Delta y_{i}$ can be regarded as a sampling measurement of $\Delta x_i$. Therefore, the problem becomes how to estimate $||\Delta x_i||_2$ from the known $||\Delta y_i||_2$, which is an ill-posed problem.
Thanks to the restricted isometry property(RIP) proposed in~\cite{candes2008restricted},
if the sampling matrix $\Phi_1$ satisfies the RIP which is the basic premise of CS, the reconstruction errors could be campled by the measurement errors with the inequality 
\begin{equation}
\setlength{\abovedisplayskip}{4pt}
\setlength{\belowdisplayskip}{4pt}
	\frac{||\Delta y||_2^{2}}{(1+\delta)} \leq ||\Delta x||_2^2 \leq \frac{||\Delta y||_2^{2}}{(1-\delta)},
\end{equation}
where $\delta \in (0, 1)$.
RIP provides a double side clamping for the $l_2$ norm of the reconstruction errors $\Delta x$. Therefore, it is reasonable to conclude that if the measurement error $||\Delta y||_2 ^ 2$ is relatively large, the corresponding reconstruction error $||\Delta x||_2 ^ 2$ is probably large as well, and vice versa.

Here, we verify the effectiveness of estimating $||\Delta x||_2 ^ 2$ from $||\Delta y||_2 ^ 2$ according to the statistical results and comparison of the quantized adaptive masks.
From the statistical results shown in Fig.~\ref{fig: dx_dy}(a), $||\Delta y||_2 ^ 2$ and $||\Delta x||_2 ^ 2$ have a nearly linear relationship. 
We also calculate the adaptive masks according to $||\Delta x||_2 ^ 2$ and $||\Delta y||_2 ^ 2$, and demonstrate the results in Fig.~\ref{fig: dx_dy}(b) and (c) respectively. Each element of the binarization mask corresponds to a region, with first $\alpha$ ($\alpha = 70\%$ in this paper) larger $||\Delta y||_2 ^ 2$ as 1 and the rest as 0.
It can be seen that the adaptive masks calculated according to $||\Delta x||_2 ^ 2$ and $||\Delta y||_2 ^ 2$ are highly consistent. 
To sum up, by calculating $||\Delta y||_2 ^ 2$, we can get which regions have larger reconstruction errors, which provides a basis for the next sampling.

\paragraph{Error based adaptive strategy}
The lower left corner of Fig.~\ref{fig: network_structure} shows the schematic diagram of the adaptive sampling module.
The error based adaptive strategy determines which image patches should be sampled at the stage \emph{i} according to the measurement error of the stage \emph{i-1} of each image patch.
Except for the stage \emph{1}, according to the adaptive mask calculated by the adaptive sampling module, the ACCSNet only samples the regions with relatively poor reconstruction quality of the previous stage. Through this adaptive and cascaded sampling mechanism, different regions of an image are sampled with different sampling rates efficiently. 
The details of Error based adaptive strategy are concluded in Alg.~\ref{alg:1}.

\begin{algorithm}
	\renewcommand{\algorithmicrequire}{\textbf{Input:}}
	\renewcommand{\algorithmicensure}{\textbf{Output:}}
	\caption{Error based adaptive strategy}
	\label{alg:1}
	\begin{algorithmic}[1]
		\REQUIRE Sampling scale factor $\alpha$, patch size $p$, samping measurement $y_1$ and sampling matrix $\Phi_1$ of the stage \emph{1}, reconstruction result $\hat{x}_{i-1}$ of the 				stage \emph{i+1}, ground truth image $x$ with scale $H\times W$
		\ENSURE Sampling measurement $y_i$ of the stage \emph{i}
		\STATE Calculate the measurement error:  \\ 
		\centerline{$\Delta y_{i-1} = y_1 - \Phi_1 \hat{x}_{i-1}$}
		\STATE Calculate the $l_2$ norm of each patch \emph{j}: \\ 
		\centerline{$V=[\|\Delta y_{i-1}^1\|_2^2,\|\Delta y_{i-1}^2\|_2^2,...,\|\Delta y_{i-1}^j\|_2^2]$}
		\STATE Calculate the number of patches not sampled of the stage \emph{i}: $N_a=(1-\alpha^{i-1}) \times 				\frac{H}{p} \times \frac{W}{p}$
		\STATE Sort $V$ from small to large: $V_{\mathrm{sorted}} = \mathrm{sort}_\uparrow(V)$ 
		\STATE Find the $N_a-th$ smallest value of V as the threshold $T$ of 	binarization: $T  = V_{\mathrm{sorted}}(N_a)$
		\STATE Binarize V based on $T$: 
		\vspace{-3mm}
		$$ B_a=\left\{
		\begin{array}{rcl}
		1  & {V > T}\\
		0  & {V \leq T}
		\end{array} \right. $$
		\vspace{-5mm}
		\STATE Expand $B_a$ to generate a pixel-wise adaptive mask $E_a$
		\STATE Sample patches based on the adaptive mask $E_a $: \\ 
		\centerline{$y_i = \Phi_i(E_a \odot x)$}
		\STATE \textbf{return} $y_i$
	\end{algorithmic}  
\end{algorithm}

\subsection{Cascaded Feature Fusion Network Structure}
As shown in Fig.~\ref{fig: network_structure}, the reconstruction model of each stage contains an initial linear layer and a reconstruction network.
The initial linear layer of the stage \emph{1} is a linear transformation, which maps the sampling results of $m_i$ channels to $p^2c$ channels through only one $1\times 1$ convolution layer, and then obtains the initial reconstruction result of the same size as the input image through rearrange operation. 
The rearrange operation includes a reshape layer and a concatenate layer. Specifically, reshape the one-dimensional reconstruction result of each image patch with the length of $p^2c$ into a three-dimensional image patch with the dimension of $p \times p\times c$, and then concatenate them into the original dimension of the image in the spatial dimension.
The initial reconstruction network of stage \emph{2} or later stages have the same $1\times 1$ convolution layer with stage \emph{1}, but different parameters. Also, there is an addition layer after the convolution layer, through the addition layer, the output of the $1\times 1$ convolution layer is added to the initial reconstruction result of the previous stage to obtain the initial reconstruction result of the current stage. 
Because of the design of the addition layer, the goal of the later stage is to reconstruct the residual between the previous stage initial reconstruction result and the ground truth image.

The reconstruction network of the stage \emph{1} adopts a symmetrical hourglass network with residual learning~\cite{he2016deep}, which contains four convolution layers (two at both ends) and three RRDBs~\cite{wang2018esrgan} (in the middle). All these convolution layers are followed by the LeakyRelu~\cite{xu2015empirical} nonlinear activation function.
The reconstruction network of the stage \emph{2} and the later stages take the initial reconstruction result of the current stage as input, and fuse the feature maps from the previous stage to output the final reconstruction result of the current stage. The reconstruction network of stage \emph{i} (\emph{i} $>$ 1) is similar to the deep reconstruction network of stage \emph{1}, but the former has three more convolution layers and three concatenate layers than the latter as shown in the lower right corner of Fig.~\ref{fig: network_structure}.

\subsection{Loss Function and Training Strategy}

For network training, we adopt $l_1$ loss to supervise the initial reconstruction result and the final reconstruction result for each stage. Given input image $x$, the learnable sampling matrix $\Phi_i$, the samping measurements of the stage \emph{i} can be denoted as $\mathcal{F}_{\Phi_i}(x)$. Therefore, for the \emph{i}-th stage, the loss function can be expressed as: 
\begin{equation}
\setlength{\abovedisplayskip}{4pt}
\setlength{\belowdisplayskip}{4pt}
\begin{aligned}
\mathcal{L}_i &= \mathcal{L}_i ^\mathrm{final}+\beta\ \mathcal{L}_i^\mathrm{init} \notag \\ 
&= \|\mathcal{N}_i^\mathrm{final}(\mathcal{F}_{\Phi_i}(x))-x\|_1+\beta\|\mathcal{N}_i^\mathrm{init}(\mathcal{F}_{\Phi_i}(x))-x\|_1 ,\notag
\end{aligned}
\end{equation}
where $\beta$, $\mathcal{N}^\mathrm{final}_i(.)$ and $\mathcal{N}^\mathrm{init}_i(.)$ denotes the loss balance hyper-parameter, the full reconstruction network and the initial reconstruction network respectively.
We train the \emph{k}-stage model gradually, which means we fix the parameters of the previous stage after we complete the training process of the previous stage and then train the next stage until we complete the training of the whole model.


\section{Experiments}
\label{sec:experiments}
\subsection{Dataset and implementation details}
We use 200 images from BSDS500 trainset~\cite{arbelaez2010contour} and 91 images from~\cite{dong2014learning} to train our model, and the validation set from BSDS500 is used for validation. Set5~\cite{bevilacqua2012low}, Set11~\cite{kulkarni2016reconnet}, Set14~\cite{zeyde2010single}, BSD68~\cite{martin2001database}, the testsets from BSDS500, General100~\cite{dong2016accelerating}, Urban100~\cite{huang2015single} and Manga109~\cite{matsui2017sketch} are used for testing. For the experiments, the Y channel of the images in YUV color space are utilized. We randomly crop $128 \times 128$ image patches from the training dataset during the training process. Batchsize is set as $2$. $\beta$ is chosen empirically as $1$. In Sec.~\ref{sec:compare_with_CS} and Sec.~\ref{sec:ablation}, we set the image patch size $p$ and sampling scale factor $\alpha$ to 8 and 0.7, respectively. Adam optimizer~\cite{kingma2014adam} is adopted with $\beta_1=0.9$, $\beta_2=0.999$ and $\epsilon=1\times10^{-8}$. We set the number of CS reconstruction stages to 4 and sampling rate to 0.05 for each stage. The proposed adaptive sampling strategy is introduce at stage 2 to 4. We train 500k iterations at each stage, the initial learning rate is set to $2\times10^{-4}$, and multiplied by $0.5$ every other 100k iterations. All the experiments are implemented on the PyTorch platform with a NVIDIA RTX 3090 GPU.

\subsection{Comparison with state-of-the-art CS algorithms}
\label{sec:compare_with_CS}
\begin{table*}
\setlength{\abovecaptionskip}{2pt}
\setlength{\belowcaptionskip}{0pt}
\centering
\caption{Performance comparison with state-of-the-art CS algorithms on Set11~\cite{kulkarni2016reconnet}, BSDS500 testset~\cite{arbelaez2010contour}, Urban100~\cite{huang2015single}, General100~\cite{dong2016accelerating} and
		Manga109~\cite{matsui2017sketch}. The best performance is highlighted in red (1st best) and blue (2nd best).}
	\scalebox{0.85}{
	\begin{tabular}{|c|c|c|c|c|c|c|c|c|c|c|c|c|c}
	\hline
{\multirow{2}[0]{*}{Datasets}} & {\multirow{2}[0]{*}{Rate}}
				& \multicolumn{2}{c|}{ReconNet ~\cite{kulkarni2016reconnet}}& \multicolumn{2}{c|}{LAPRAN ~\cite{xu2018lapran}} & \multicolumn{2}{c|}{ISTA-Net+ ~\cite{zhang2018ista}}
				& \multicolumn{2}{c|}{SCSNet ~\cite{shi2019scalable}} & \multicolumn{2}{c|}{MAC-Net ~\cite{chen2020learning}}& \multicolumn{2}{c|}{Ours}\\
	\cline{3-14}
				& \multicolumn{1}{c|}{} & \multicolumn{1}{c|}{PSNR} & \multicolumn{1}{c|}{SSIM} & \multicolumn{1}{c|}{PSNR} & \multicolumn{1}{c|}{SSIM} 
				& \multicolumn{1}{c|}{PSNR} & \multicolumn{1}{c|}{SSIM} & \multicolumn{1}{c|}{PSNR} & \multicolumn{1}{c|}{SSIM} 
				& \multicolumn{1}{c|}{PSNR} & \multicolumn{1}{c|}{SSIM }& \multicolumn{1}{c|}{PSNR} & \multicolumn{1}{c|}{SSIM}\\
	\hline
{\multirow{4}[0]{*}{Set11}} 
				& \multicolumn{1}{c|}{0.05} & \multicolumn{1}{c|}{22.24} & \multicolumn{1}{c|}{0.6352} & \multicolumn{1}{c|}{21.58} & \multicolumn{1}{c|}{0.5913} & \multicolumn{1}{c|}{22.70}
				& \multicolumn{1}{c|}{0.6661} & \multicolumn{1}{c|}{\blue{24.65}} & \multicolumn{1}{c|}{\blue{0.7290}} & \multicolumn{1}{c|}{22.77} & \multicolumn{1}{c|}{0.7067} & \multicolumn{1}{c|}{\red{26.61}}& \multicolumn{1}{c|}{\red{0.8162}}\\
				& \multicolumn{1}{c|}{0.1} & \multicolumn{1}{c|}{24.80} & \multicolumn{1}{c|}{0.7284} & \multicolumn{1}{c|}{23.78} & \multicolumn{1}{c|}{0.6893} & \multicolumn{1}{c|}{26.53}
				& \multicolumn{1}{c|}{0.7992} & \multicolumn{1}{c|}{\blue{28.16}} & \multicolumn{1}{c|}{\blue{0.8506}} & \multicolumn{1}{c|}{25.91} & \multicolumn{1}{c|}{0.8042} & \multicolumn{1}{c|}{\red{29.76}}& \multicolumn{1}{c|}{\red{0.8847}}\\
				& \multicolumn{1}{c|}{0.15} & \multicolumn{1}{c|}{25.55} & \multicolumn{1}{c|}{0.7733} & \multicolumn{1}{c|}{24.27} & \multicolumn{1}{c|}{0.6897} & \multicolumn{1}{c|}{28.67}
				& \multicolumn{1}{c|}{0.8541} & \multicolumn{1}{c|}{\blue{29.75}} & \multicolumn{1}{c|}{\blue{0.8875}} & \multicolumn{1}{c|}{27.56} & \multicolumn{1}{c|}{0.8478} & \multicolumn{1}{c|}{\red{31.86}}& \multicolumn{1}{c|}{\red{0.9139}}\\
				& \multicolumn{1}{c|}{0.2} & \multicolumn{1}{c|}{27.39} & \multicolumn{1}{c|}{0.8110} & \multicolumn{1}{c|}{26.04} & \multicolumn{1}{c|}{0.7727} & \multicolumn{1}{c|}{30.56}
				& \multicolumn{1}{c|}{0.8900} & \multicolumn{1}{c|}{\blue{31.30}} & \multicolumn{1}{c|}{\blue{0.9132}} & \multicolumn{1}{c|}{29.08} & \multicolumn{1}{c|}{0.8797} & \multicolumn{1}{c|}{\red{33.61}}& \multicolumn{1}{c|}{\red{0.9309}}\\
	\hline
 {\multirow{4}[0]{*}{BSDS500}} 
				& \multicolumn{1}{c|}{0.05} & \multicolumn{1}{c|}{24.07} & \multicolumn{1}{c|}{0.6189} & \multicolumn{1}{c|}{23.58} & \multicolumn{1}{c|}{0.5812} & \multicolumn{1}{c|}{24.18}
				& \multicolumn{1}{c|}{0.6289} & \multicolumn{1}{c|}{\red{26.77}} & \multicolumn{1}{c|}{\blue{0.6972}} & \multicolumn{1}{c|}{23.77} & \multicolumn{1}{c|}{0.6489} & \multicolumn{1}{c|}{\blue{26.72}}& \multicolumn{1}{c|}{\red{0.7310}}\\
				& \multicolumn{1}{c|}{0.1} & \multicolumn{1}{c|}{25.79} & \multicolumn{1}{c|}{0.6881} & \multicolumn{1}{c|}{25.11} & \multicolumn{1}{c|}{0.6517} & \multicolumn{1}{c|}{26.50}
				& \multicolumn{1}{c|}{0.7306} & \multicolumn{1}{c|}{\blue{28.57}} & \multicolumn{1}{c|}{\blue{0.7844}} & \multicolumn{1}{c|}{25.97} & \multicolumn{1}{c|}{0.7341} & \multicolumn{1}{c|}{\red{29.33}}& \multicolumn{1}{c|}{\red{0.8292}}\\
				& \multicolumn{1}{c|}{0.15} & \multicolumn{1}{c|}{26.39} & \multicolumn{1}{c|}{0.7337} & \multicolumn{1}{c|}{25.38} & \multicolumn{1}{c|}{0.6500} & \multicolumn{1}{c|}{28.10}
				& \multicolumn{1}{c|}{0.7874} & \multicolumn{1}{c|}{\blue{29.79}} & \multicolumn{1}{c|}{\blue{0.8509}} & \multicolumn{1}{c|}{27.06} & \multicolumn{1}{c|}{0.7800} & \multicolumn{1}{c|}{\red{31.11}}& \multicolumn{1}{c|}{\red{0.8713}}\\
				& \multicolumn{1}{c|}{0.2} & \multicolumn{1}{c|}{27.63} & \multicolumn{1}{c|}{0.7665} & \multicolumn{1}{c|}{26.77} & \multicolumn{1}{c|}{0.7271} & \multicolumn{1}{c|}{29.38}
				& \multicolumn{1}{c|}{0.8295} & \multicolumn{1}{c|}{\blue{31.10}} & \multicolumn{1}{c|}{\blue{0.8731}} & \multicolumn{1}{c|}{28.35} & \multicolumn{1}{c|}{0.8210} & \multicolumn{1}{c|}{\red{32.67}}& \multicolumn{1}{c|}{\red{0.8976}}\\
	\hline
{\multirow{4}[0]{*}{Urban100}} 
				& \multicolumn{1}{c|}{0.05} & \multicolumn{1}{c|}{21.36} & \multicolumn{1}{c|}{0.5761} & \multicolumn{1}{c|}{20.85} & \multicolumn{1}{c|}{0.5331} & \multicolumn{1}{c|}{21.89}
				& \multicolumn{1}{c|}{0.6117} & \multicolumn{1}{c|}{\blue{23.28}} & \multicolumn{1}{c|}{\blue{0.6514}} & \multicolumn{1}{c|}{21.56} & \multicolumn{1}{c|}{0.6347} & \multicolumn{1}{c|}{\red{24.89}}& \multicolumn{1}{c|}{\red{0.7444}}\\
				& \multicolumn{1}{c|}{0.1} & \multicolumn{1}{c|}{23.31} & \multicolumn{1}{c|}{0.6622} & \multicolumn{1}{c|}{22.51} & \multicolumn{1}{c|}{0.6171} & \multicolumn{1}{c|}{24.87}
				& \multicolumn{1}{c|}{0.7392} & \multicolumn{1}{c|}{\blue{26.03}} & \multicolumn{1}{c|}{\blue{0.7873}} & \multicolumn{1}{c|}{24.23} & \multicolumn{1}{c|}{0.7404} & \multicolumn{1}{c|}{\red{27.80}}& \multicolumn{1}{c|}{\red{0.8422}}\\
				& \multicolumn{1}{c|}{0.15} & \multicolumn{1}{c|}{24.18} & \multicolumn{1}{c|}{0.7146} & \multicolumn{1}{c|}{23.03} & \multicolumn{1}{c|}{0.6294} & \multicolumn{1}{c|}{26.88}
				& \multicolumn{1}{c|}{0.8054} & \multicolumn{1}{c|}{\blue{27.36}} & \multicolumn{1}{c|}{\blue{0.8348}} & \multicolumn{1}{c|}{25.54} & \multicolumn{1}{c|}{0.7896} & \multicolumn{1}{c|}{\red{29.62}}& \multicolumn{1}{c|}{\red{0.8793}}\\
				& \multicolumn{1}{c|}{0.2} & \multicolumn{1}{c|}{25.47} & \multicolumn{1}{c|}{0.7535} & \multicolumn{1}{c|}{24.28} & \multicolumn{1}{c|}{0.7027} & \multicolumn{1}{c|}{28.35}
				& \multicolumn{1}{c|}{0.8499} & \multicolumn{1}{c|}{\blue{28.60}} & \multicolumn{1}{c|}{\blue{0.8700}} & \multicolumn{1}{c|}{27.01} & \multicolumn{1}{c|}{0.8326} & \multicolumn{1}{c|}{\red{31.08}}& \multicolumn{1}{c|}{\red{0.9009}}\\
	\hline
{\multirow{4}[0]{*}{General100}} 
				& \multicolumn{1}{c|}{0.05} & \multicolumn{1}{c|}{25.28} & \multicolumn{1}{c|}{0.6990} & \multicolumn{1}{c|}{24.66} & \multicolumn{1}{c|}{0.6600} & \multicolumn{1}{c|}{26.11}
				& \multicolumn{1}{c|}{0.7182} & \multicolumn{1}{c|}{\blue{28.06}} & \multicolumn{1}{c|}{\blue{0.7702}} & \multicolumn{1}{c|}{26.42} & \multicolumn{1}{c|}{0.7482} & \multicolumn{1}{c|}{\red{30.07}}& \multicolumn{1}{c|}{\red{0.8308}}\\
				& \multicolumn{1}{c|}{0.1} & \multicolumn{1}{c|}{27.91} & \multicolumn{1}{c|}{0.7702} & \multicolumn{1}{c|}{26.81} & \multicolumn{1}{c|}{0.7333} & \multicolumn{1}{c|}{29.82}
				& \multicolumn{1}{c|}{0.8218} & \multicolumn{1}{c|}{\blue{31.88}} & \multicolumn{1}{c|}{\blue{0.8754}} & \multicolumn{1}{c|}{29.54} & \multicolumn{1}{c|}{0.8276} & \multicolumn{1}{c|}{\red{33.54}}& \multicolumn{1}{c|}{\red{0.9013}}\\
				& \multicolumn{1}{c|}{0.15} & \multicolumn{1}{c|}{27.90} & \multicolumn{1}{c|}{0.8094} & \multicolumn{1}{c|}{26.91} & \multicolumn{1}{c|}{0.7254} & \multicolumn{1}{c|}{31.95}
				& \multicolumn{1}{c|}{0.8683} & \multicolumn{1}{c|}{\blue{33.63}} & \multicolumn{1}{c|}{\blue{0.9107}} & \multicolumn{1}{c|}{31.18} & \multicolumn{1}{c|}{0.8658} & \multicolumn{1}{c|}{\red{35.64}}& \multicolumn{1}{c|}{\red{0.9267}}\\
				& \multicolumn{1}{c|}{0.2} & \multicolumn{1}{c|}{29.91} & \multicolumn{1}{c|}{0.8375} & \multicolumn{1}{c|}{29.07} & \multicolumn{1}{c|}{0.8047} & \multicolumn{1}{c|}{33.66}
				& \multicolumn{1}{c|}{0.8996} & \multicolumn{1}{c|}{\blue{35.23}} & \multicolumn{1}{c|}{\blue{0.9341}} & \multicolumn{1}{c|}{32.93} & \multicolumn{1}{c|}{0.8968} & \multicolumn{1}{c|}{\red{37.23}}& \multicolumn{1}{c|}{\red{0.9418}}\\
	\hline
{\multirow{4}[0]{*}{Manga109}} 
				& \multicolumn{1}{c|}{0.05} & \multicolumn{1}{c|}{23.04} & \multicolumn{1}{c|}{0.7025} & \multicolumn{1}{c|}{22.24} & \multicolumn{1}{c|}{0.6570} & \multicolumn{1}{c|}{24.17}
				& \multicolumn{1}{c|}{0.7518} & \multicolumn{1}{c|}{\blue{25.10}} & \multicolumn{1}{c|}{0.7520} & \multicolumn{1}{c|}{23.77} & \multicolumn{1}{c|}{\blue{0.7786}} & \multicolumn{1}{c|}{\red{28.26}}& \multicolumn{1}{c|}{\red{0.8731}}\\
				& \multicolumn{1}{c|}{0.1} & \multicolumn{1}{c|}{25.96} & \multicolumn{1}{c|}{0.7830} & \multicolumn{1}{c|}{24.64} & \multicolumn{1}{c|}{0.7418} & \multicolumn{1}{c|}{28.84}
				& \multicolumn{1}{c|}{0.8675} & \multicolumn{1}{c|}{\blue{30.17}} & \multicolumn{1}{c|}{\blue{0.8968}} & \multicolumn{1}{c|}{27.67} & \multicolumn{1}{c|}{0.8675} & \multicolumn{1}{c|}{\red{33.27}} & \multicolumn{1}{c|}{\red{0.9384}}\\
				& \multicolumn{1}{c|}{0.15} & \multicolumn{1}{c|}{26.65} & \multicolumn{1}{c|}{0.8254} & \multicolumn{1}{c|}{25.14} & \multicolumn{1}{c|}{0.7414} & \multicolumn{1}{c|}{31.69}
				& \multicolumn{1}{c|}{0.9099} & \multicolumn{1}{c|}{\blue{32.36}} & \multicolumn{1}{c|}{\blue{0.9292}} & \multicolumn{1}{c|}{29.83} & \multicolumn{1}{c|}{0.9039} & \multicolumn{1}{c|}{\red{36.45}}& \multicolumn{1}{c|}{\red{0.9567}}\\
				& \multicolumn{1}{c|}{0.2} & \multicolumn{1}{c|}{28.96} & \multicolumn{1}{c|}{0.8553} & \multicolumn{1}{c|}{27.33} & \multicolumn{1}{c|}{0.8199} & \multicolumn{1}{c|}{33.90}
				& \multicolumn{1}{c|}{0.9347} & \multicolumn{1}{c|}{\blue{34.48}} & \multicolumn{1}{c|}{\blue{0.9508}} & \multicolumn{1}{c|}{31.92} & \multicolumn{1}{c|}{0.9294} & \multicolumn{1}{c|}{\red{38.70}} & \multicolumn{1}{c|}{\red{0.9652}}\\
	\hline
	\end{tabular}
			}
	\label{tab:compare_with_sota}
\vspace{-3mm}
\end{table*}

In this section, we compare with other CS algorithms proposed in recent years, i.e. ReconNet~\cite{kulkarni2016reconnet}, LAPRAN~\cite{xu2018lapran}, ISTA-Net+~\cite{zhang2018ista}, SCSNet~\cite{shi2019scalable} and MAC-Net~\cite{chen2020learning}.
For comparison, we use the code provided by the author and the settings of training and testing are chosen according to the original paper.
The quantitative results comparison is shown in Tab.~\ref{tab:compare_with_sota}.
It can be seen that the reconstruction performance of the proposed adaptive CS algorithm are largely improved than the existing CS algorithms in both PSNR and SSIM, especially at the sampling rates of 0.1, 0.15 and 0.2, when the ASM is introduced. 

\begin{figure*}
\setlength{\abovecaptionskip}{2pt}
\setlength{\belowcaptionskip}{-3mm}
\centering
\scalebox{0.95}{
\includegraphics[width=\linewidth]{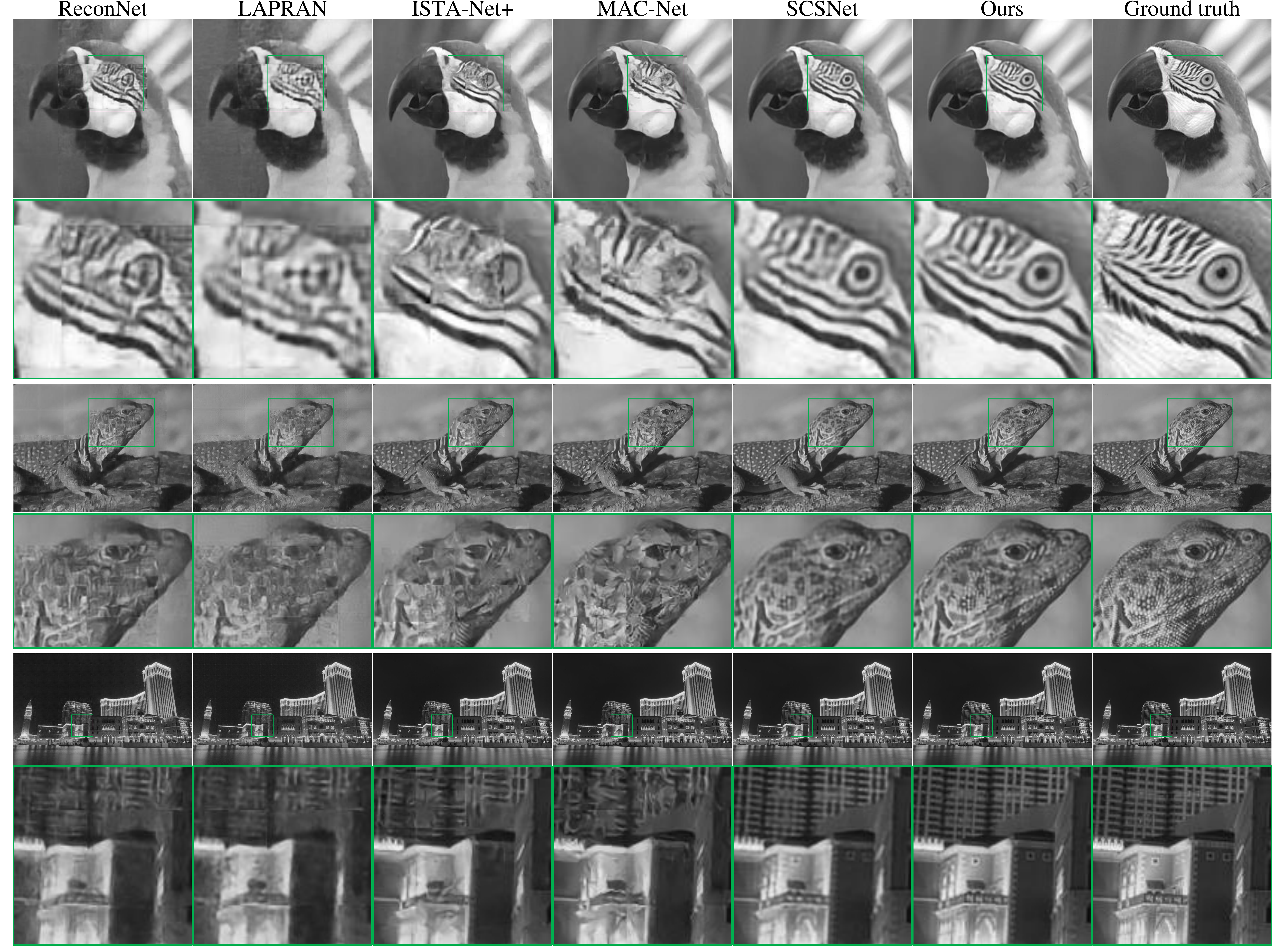}
}
	\caption{Visual comparison with the state-of-the-art CS algorithms. Top rows: \emph{Parrots} from Set11~\cite{kulkarni2016reconnet} with sampling rate = 0.1, middle rows: \emph{41096} from BSDS500 testset~\cite{arbelaez2010contour} with sampling rate = 0.15, bottom rows: \emph{img\_085} from Urban100~\cite{huang2015single} with sampling rate = 0.2.}
	\label{fig: compare_with_sota}
\end{figure*}

Fig.~\ref{fig: compare_with_sota} shows the reconstruction results with the sampling rates of 0.1, 0.15 and 0.2 respectively. 
It can be seen that at a sampling rate of 0.1, our method can reconstruct the head of the parrot more accurately, with much higher visual quality. With the proposed ASM, different sampling rates are allocated for different regions according to the reconstruction error of different regions. During the adaptive sampling process, since the reconstruction error of the head area is high, the sampling rate of that region is then adaptively increased and thus enables to achieve better reconstruction results at the same sampling rates. The same comparison results are shown at the lizard and night scene at the sampling rates of 0.15 and 0.2 respectively in Fig.~\ref{fig: compare_with_sota}. As are shown, the reconstruction accuracy of the structures on the head of the lizard and the appearance of the house are largely improved, with the same overall CS sampling rates with the other methods. 

\subsection{Ablation study}
\label{sec:ablation}

\begin{figure}[htbp]
\setlength{\belowcaptionskip}{-7mm}
\centering
\includegraphics[width=\linewidth]{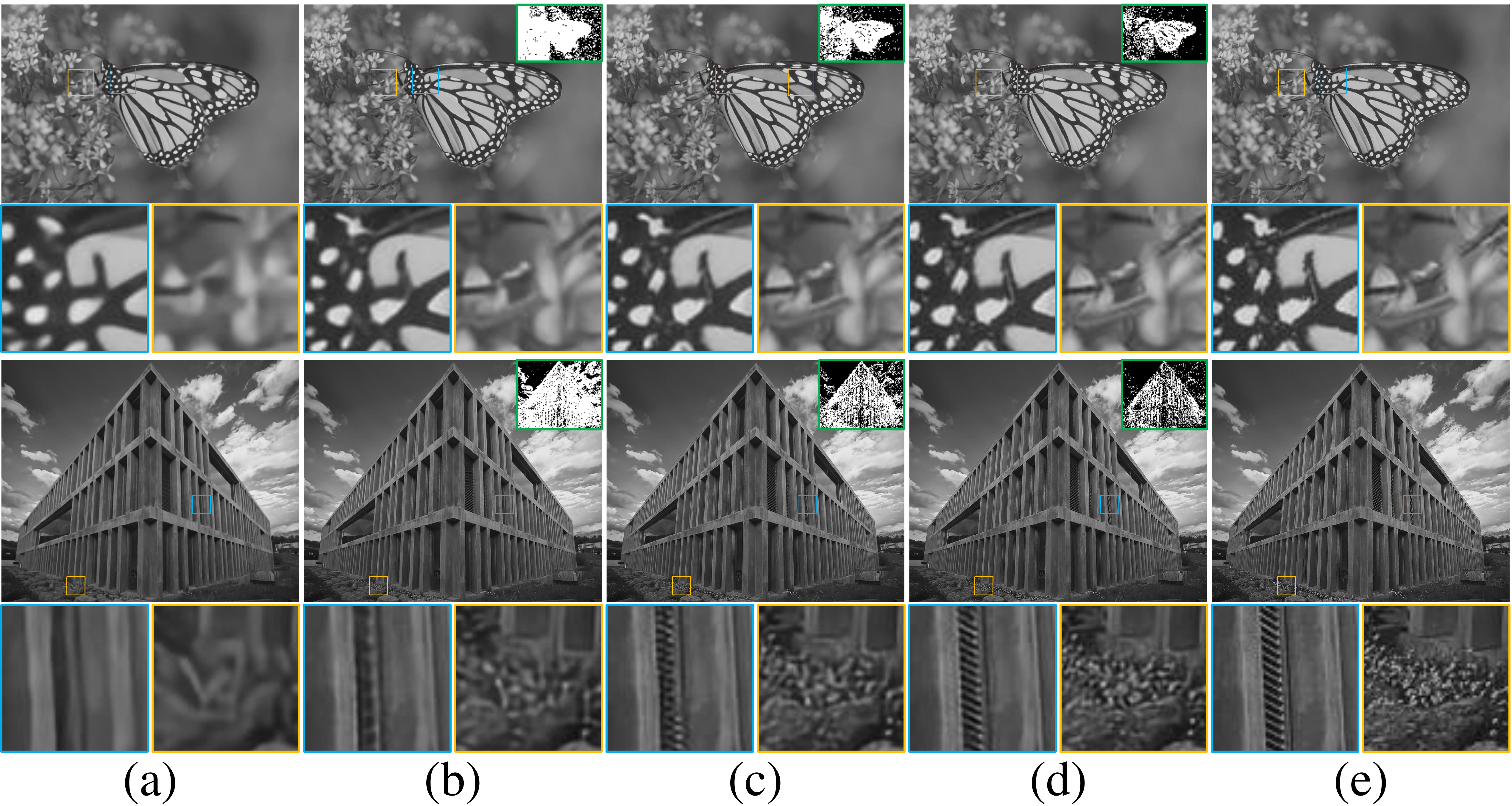}
\caption{Reconstruction results of four stages, the corresponding adaptive mask is displayed in the top right of the image, which is narrowed and framed with a green box. (a) Stage 1, sampling rate = 0.05, (b) stage 2, sampling rate = 0.1, (c) stage 3, sampling rate = 0.15, (d) stage 4, sampling rate = 0.2, (e) ground truth.}
	\label{fig: Amap_residual_res_difCascade}
\end{figure}

\begin{figure}[htbp]
\setlength{\belowcaptionskip}{-7mm}
\centering
\includegraphics[width=\linewidth]{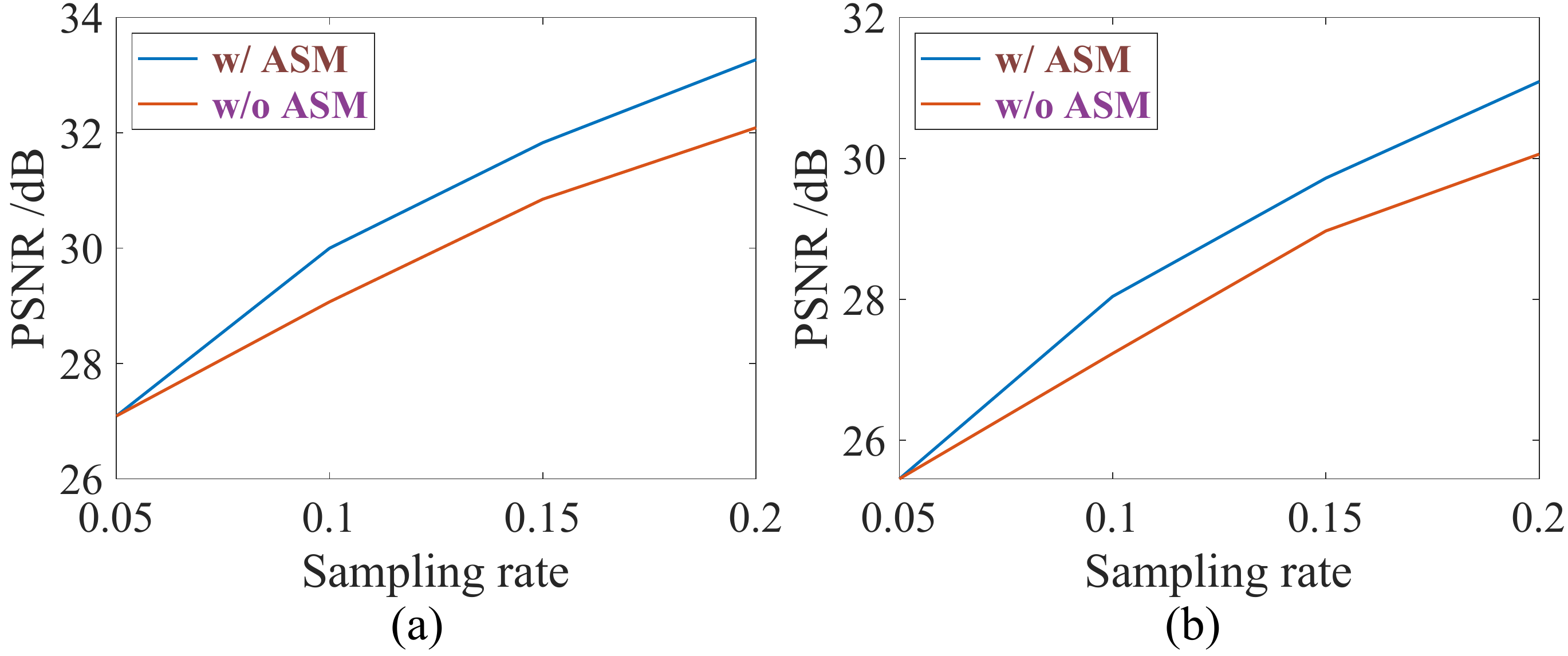}
\caption{Ablation study of ASM on (a) Set14~\cite{zeyde2010single}, (b) BSD68~\cite{martin2001database}. w/o and w/ denote without and with respectively.}
	\label{fig: ablation_our_attention}
\end{figure}

\begin{figure}[htbp]
\setlength{\belowcaptionskip}{-3mm}
\centering
\includegraphics[width=\linewidth]{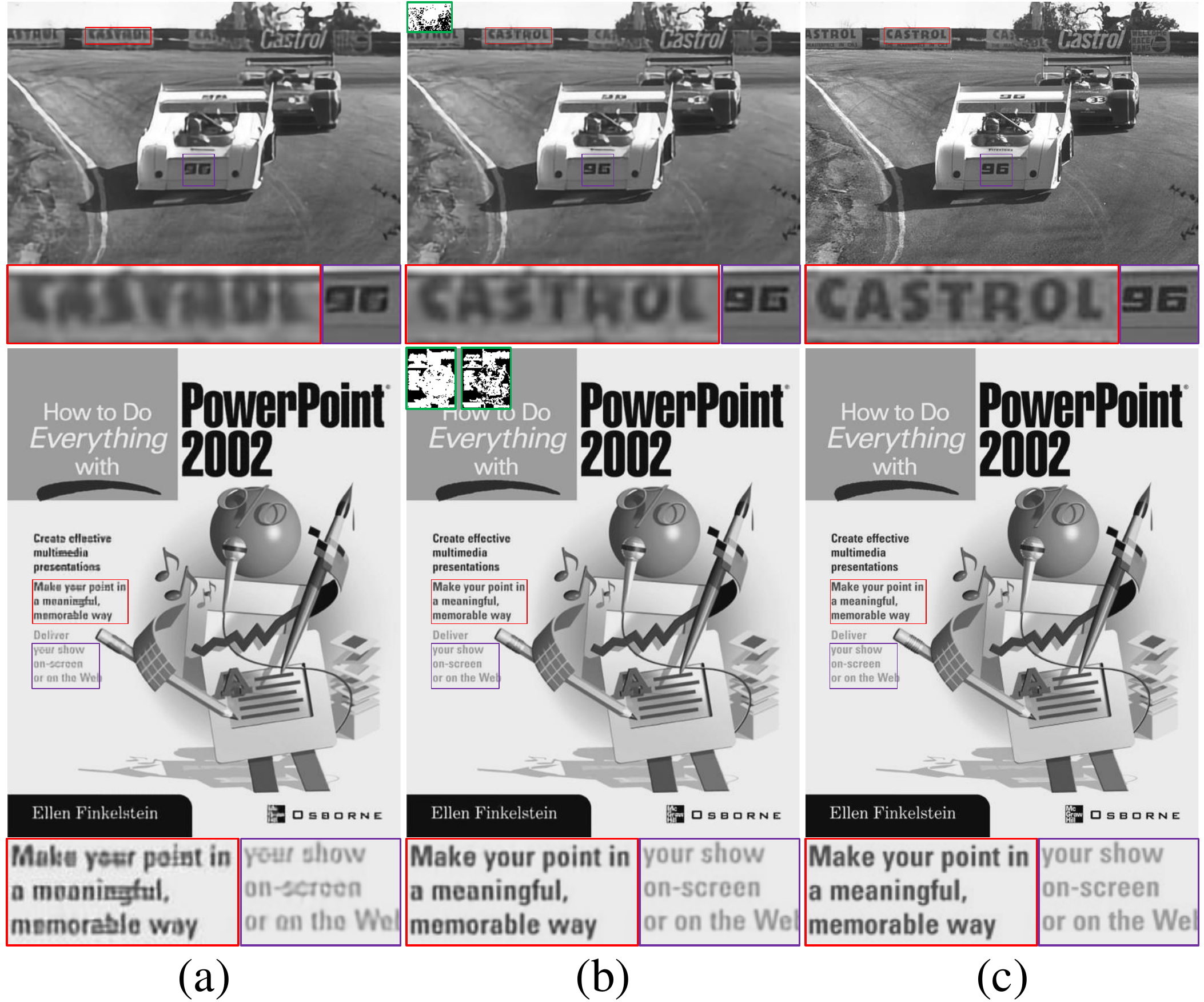}
\caption{Ablation study of ASM with sampling rate = 0.1 on \emph{test040} (1st and 2nd row) from BSD68~\cite{martin2001database} and sampling rate = 0.15 on \emph{PPT3} (3rd and 4th row) from Set14~\cite{zeyde2010single}. The adaptive masks generated by ASM are placed in the upper left corner of the images of the 2nd column and framed with a green box. (a) Without ASM,  (b) with ASM, (c) ground truth.}
	\label{fig: img_ablation_our_attention_set14}
\end{figure}

\paragraph{Effectiveness of ASM} Fig.~\ref{fig: Amap_residual_res_difCascade} shows the reconstruction results of different stages and the corresponding binary adaptive masks. The binary adaptive masks adaptively focus on the regions with more reconstruction errors, and the reconstruction results are gradually improved.
We implement an ablation study to verify the effectiveness of the proposed ASM.
Fig.~\ref{fig: ablation_our_attention} shows the PSNR of our algorithm in the reconstruction results with the ASM and without the ASM.
Except for stage 1, the proposed ASM can bring about 1dB improvement in PSNR.
From the visual reconstruction results shown in Fig.~\ref{fig: img_ablation_our_attention_set14}, the model with the ASM can reconstruct more accurate image details, while there are serious artifacts in the reconstruction results without the ASM at the same overall sampling rate.

\begin{table}[htbp]
\setlength{\abovecaptionskip}{2pt}
\setlength{\belowcaptionskip}{0pt}
\centering
\caption{Quantitative reconstruction results with and without feature fusion on Set5~\cite{bevilacqua2012low} and BSDS500 testset~\cite{arbelaez2010contour}.}
	\scalebox{0.9}{
	\begin{tabular}{c|c|c|c|c|c}
	\hline
{\multirow{2}[0]{*}{Datasets}} & {\multirow{2}[0]{*}{Rate}}
				& \multicolumn{2}{c|}{w/o feature fusion}& \multicolumn{2}{c}{w/ feature fusion} \\
	\cline{3-6}
				& \multicolumn{1}{c|}{} & \multicolumn{1}{c|}{PSNR} & \multicolumn{1}{c|}{SSIM} & \multicolumn{1}{c|}{PSNR} & \multicolumn{1}{c}{SSIM} \\
	\hline
{\multirow{4}[0]{*}{Set5}} 
				& \multicolumn{1}{c|}{0.05} & \multicolumn{1}{c|}{30.01} & \multicolumn{1}{c|}{0.8602} & \multicolumn{1}{c|}{30.01} & \multicolumn{1}{c}{0.8602} \\
				& \multicolumn{1}{c|}{0.1} & \multicolumn{1}{c|}{33.95} & \multicolumn{1}{c|}{0.9210} & \multicolumn{1}{c|}{34.04} & \multicolumn{1}{c}{0.9216} \\
				& \multicolumn{1}{c|}{0.15} & \multicolumn{1}{c|}{35.58} & \multicolumn{1}{c|}{0.9354} & \multicolumn{1}{c|}{35.89} & \multicolumn{1}{c}{0.9380} \\
				& \multicolumn{1}{c|}{0.2} & \multicolumn{1}{c|}{36.70} & \multicolumn{1}{c|}{0.9438} & \multicolumn{1}{c|}{37.19} & \multicolumn{1}{c}{0.9477} \\
	\hline
{\multirow{4}[0]{*}{BSDS500}} 
				& \multicolumn{1}{c|}{0.05} & \multicolumn{1}{c|}{26.72} & \multicolumn{1}{c|}{0.7310} & \multicolumn{1}{c|}{26.72} & \multicolumn{1}{c}{0.7310} \\
				& \multicolumn{1}{c|}{0.1} & \multicolumn{1}{c|}{29.25} & \multicolumn{1}{c|}{0.8276} & \multicolumn{1}{c|}{29.33} & \multicolumn{1}{c}{0.8292} \\
				& \multicolumn{1}{c|}{0.15} & \multicolumn{1}{c|}{30.90} & \multicolumn{1}{c|}{0.8693} & \multicolumn{1}{c|}{31.11} & \multicolumn{1}{c}{0.8713} \\
				& \multicolumn{1}{c|}{0.2} & \multicolumn{1}{c|}{32.28} & \multicolumn{1}{c|}{0.8920} & \multicolumn{1}{c|}{32.67} & \multicolumn{1}{c}{0.8976} \\
	\hline
	\end{tabular}
			}
	\label{tab:ablation_feafuse}
\vspace{-5mm}
\end{table}

\begin{figure}[htbp]
\setlength{\belowcaptionskip}{-3mm}
\centering
\includegraphics[width=\linewidth]{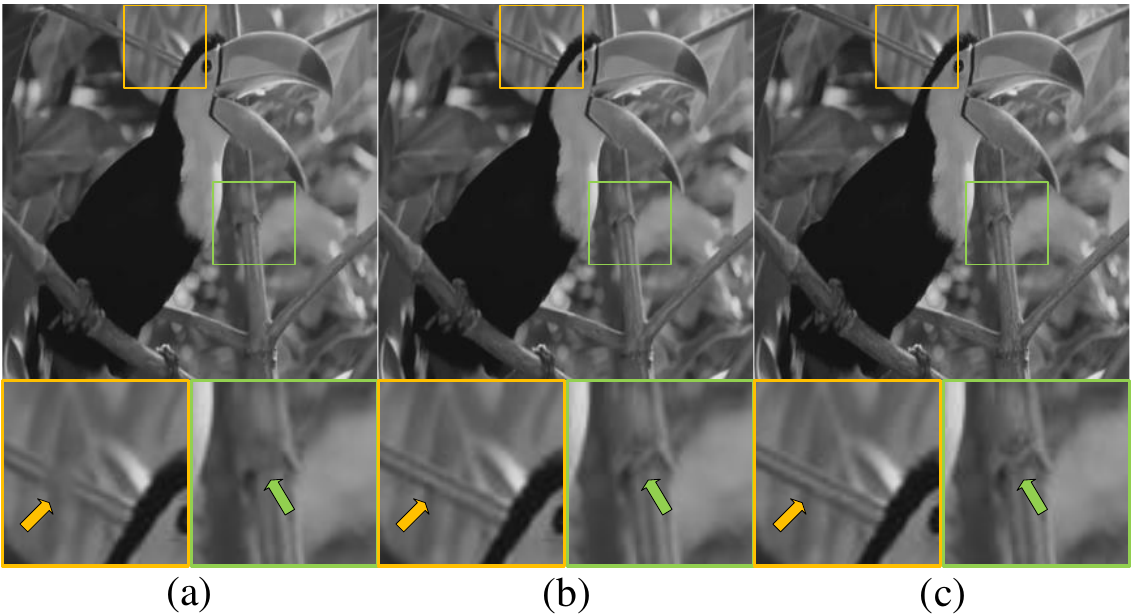}
\caption{Ablation study of feature fusion on \emph{bird} from Set5~\cite{bevilacqua2012low} with sampling rate = 0.15. (a) Without feature fusion, (b) with feature fusion, (c) ground truth.}
	\label{fig: ablation_feafuse_Set5}
\end{figure}

\paragraph{Effectiveness of feature fusion} We propose cascaded feature-fusion connections between different stages to fuse the features of the previous stage into the reconstruction network of the current stage and assist the proposed ASM in highly-efficient image reconstruction.
Tab.~\ref{tab:ablation_feafuse} shows the quantitative results with and without feature fusion connections between stages. Note that the improvement with cascaded feature fusion starts from stage 2, the improvement of introducing the cascaded feature fusion starts from stage 2. From comparison results, we can see that the quantitative results of the reconstruction results of our method with cascaded feature fusion are better than those without cascaded feature fusion among stages. 
Fig.~\ref{fig: ablation_feafuse_Set5} shows the visual reconstruction results of different models. The model with cascaded feature fusion can reconstruct elegantly higher quality and better visual results.
As compared, the cascaded feature fusion enables the information of the previous stage to be transmitted to the later stage, and can make full use of the information among stages for further reconstruction quality improvement.

\subsection{Exploration of model hyper-parameters}

\begin{table}[htbp]
\setlength{\abovecaptionskip}{2pt}
\setlength{\belowcaptionskip}{0pt}
\centering
\caption{Quantitative results with different patch sizes on Set14~\cite{zeyde2010single} and General100~\cite{dong2016accelerating}.}
	\scalebox{0.72}{
	\begin{threeparttable}
	\begin{tabular}{c|c|c|c|c|c|c|c}
	\hline
{\multirow{2}[0]{*}{Datasets}} & {\multirow{2}[0]{*}{Rate}} & \multicolumn{2}{c|}{patch size = 8} 
				& \multicolumn{2}{c|}{patch size = 16}& \multicolumn{2}{c}{patch size = 32} \\
	\cline{3-8}
				& \multicolumn{1}{c|}{} & \multicolumn{1}{c|}{PSNR} & \multicolumn{1}{c|}{SSIM} & \multicolumn{1}{c|}{PSNR} & \multicolumn{1}{c|}{SSIM} 
				& \multicolumn{1}{c|}{PSNR} & \multicolumn{1}{c}{SSIM}\\
	\hline
{\multirow{4}[0]{*}{Set14}} 
				& \multicolumn{1}{c|}{0.05\tnote{*}} & \multicolumn{1}{c|}{27.09} & \multicolumn{1}{c|}{\blue{0.7467}} & \multicolumn{1}{c|}{\red{27.30}} & \multicolumn{1}{c|}{\red{0.7525}} 
				& \multicolumn{1}{c|}{\blue{27.16}} & \multicolumn{1}{c}{0.7449}\\
				& \multicolumn{1}{c|}{0.1} & \multicolumn{1}{c|}{\red{30.00}} & \multicolumn{1}{c|}{\red{0.8343}} & \multicolumn{1}{c|}{\blue{29.99}} & \multicolumn{1}{c|}{\blue{0.8331}} 
				& \multicolumn{1}{c|}{29.84} & \multicolumn{1}{c}{0.8257}\\
				& \multicolumn{1}{c|}{0.15} & \multicolumn{1}{c|}{\blue{31.83}} & \multicolumn{1}{c|}{\red{0.8700}} & \multicolumn{1}{c|}{\red{31.86}} & \multicolumn{1}{c|}{\blue{0.8676}}
				& \multicolumn{1}{c|}{31.28} & \multicolumn{1}{c}{0.8609} \\
				& \multicolumn{1}{c|}{0.2} & \multicolumn{1}{c|}{\red{33.27}} & \multicolumn{1}{c|}{\red{0.8918}} & \multicolumn{1}{c|}{\blue{32.76}} & \multicolumn{1}{c|}{\blue{0.8861}}
				& \multicolumn{1}{c|}{32.35} & \multicolumn{1}{c}{0.8805}\\
	\hline
{\multirow{4}[0]{*}{General100}} 
				& \multicolumn{1}{c|}{0.05\tnote{*}} & \multicolumn{1}{c|}{29.81} & \multicolumn{1}{c|}{\blue{0.8261}} & \multicolumn{1}{c|}					{\red{30.08}} & \multicolumn{1}{c|}{\red{0.8315}} & \multicolumn{1}{c|}{\blue{29.94}} & \multicolumn{1}{c}{0.8251} \\
				& \multicolumn{1}{c|}{0.1} & \multicolumn{1}{c|}{\red{33.54}} & \multicolumn{1}{c|}{\red{0.9013}} & \multicolumn{1}{c|}						{\blue{33.44}} & \multicolumn{1}{c|}{\blue{0.8977}} & \multicolumn{1}{c|}{33.21} & \multicolumn{1}{c}{0.8917}\\
				& \multicolumn{1}{c|}{0.15} & \multicolumn{1}{c|}{\red{35.64}} & \multicolumn{1}{c|}{\red{0.9267}} & \multicolumn{1}{c|}							{\blue{35.52}} & \multicolumn{1}{c|}{\blue{0.9238}} & \multicolumn{1}{c|}{34.76} & \multicolumn{1}{c}{0.9172}\\
				& \multicolumn{1}{c|}{0.2} & \multicolumn{1}{c|}{\red{37.23}} & \multicolumn{1}{c|}{\red{0.9418}} & \multicolumn{1}{c|}						{\blue{36.52}} & \multicolumn{1}{c|}{\blue{0.9371}} 
				& \multicolumn{1}{c|}{35.93} & \multicolumn{1}{c}{0.9319}\\
	\hline
	\end{tabular}
	\begin{tablenotes}
       \footnotesize
       \item[*] When the sampling rate is 0.05, one of the main reasons for the difference of reconstruction results is the error introduced when calculating the number of corresponding sampling points by rounding method. For example, when the patch size is 8, the number of sampling points of each patch should be calculated as $\lfloor 0.05 \times 8 ^ 2 \rceil = 3$, and the corresponding real sampling rate should be $3 / 64 = 0.0469$. Similarly, when the patch size is 16, the corresponding real sampling rate should be 0.0508. 
     \end{tablenotes}
   \end{threeparttable}
			}
	\label{tab:dif_patchsize}
\vspace{-5mm}
\end{table}

\paragraph{Exploration of different image patch size} The sampling model proposed in this paper divides the image into non-overlapping small patches and samples different patches respectively. In the sampling model of stage 2 and later stages, the adaptive module determines whether different patches need to be sampled or not. If the image is divided with smaller patch size, the proposed ASM can determine the adaptive sampling distribution with finer scale, which could help to reconstruct the whole image with higher quality. In order to choose a proper patch size, we conduct the patch-size analysis experiments and the quantitative results are concluded in Tab.~\ref{tab:dif_patchsize}. As shown, the overall performance at patch size 8, 16, 32 are slightly different and the difference increases as the sampling rate increases. According to the quantitative comparison results, we choose 8 as the patch size in our experiments.

\begin{table}[htbp]
\setlength{\abovecaptionskip}{2pt}
\setlength{\belowcaptionskip}{0pt}
\centering
\caption{Average quantitative results of different sampling rates at different sampling scale factors $\alpha$ on Set5 ~\cite{zeyde2010single} and BSDS500 ~\cite{dong2016accelerating}.}
	\scalebox{0.75}{
	\begin{tabular}{c|c|c|c|c}
	\hline
{\multirow{2}[0]{*}{Datasets}} & \multicolumn{1}{c|}{$\alpha$ = 0.5} 
				& \multicolumn{1}{c|}{$\alpha$ = 0.6}& \multicolumn{1}{c|}{$\alpha$ = 0.7}& \multicolumn{1}{c}{$\alpha$ = 0.8} \\
	\cline{2-5}
				& \multicolumn{1}{c|}{PSNR/SSIM} & \multicolumn{1}{c|}{PSNR/SSIM} & \multicolumn{1}{c|}{PSNR/SSIM} & \multicolumn{1}{c}{PSNR/SSIM}\\
	\hline
{\multirow{1}[0]{*}{Set5}} 
				& \multicolumn{1}{c|}{34.20/0.9144} & \multicolumn{1}{c|}{34.27/0.9164} & \multicolumn{1}{c|}{34.28/0.9169} & \multicolumn{1}{c}{34.08/0.9164}\\
	\hline
{\multirow{1}[0]{*}{BSDS500}} 
				& \multicolumn{1}{c|}{30.01/0.8275} & \multicolumn{1}{c|}{30.02/0.8310} & \multicolumn{1}{c|}{30.02/0.8323} & \multicolumn{1}{c}{29.72/0.8307}\\
	\hline
	\end{tabular}
			}
	\label{tab:dif_sampling_scale_factor}
\vspace{-5mm}
\end{table}

\paragraph{Exploration of sampling scale factor $\alpha$} 
Different sampling scale factors $\alpha$ will affect the proportion of 1 in the adaptive mask, that is, the number of sampling regions of each adaptive sampling stage. 
A larger sampling scale factor $\alpha$ can allocate more regions of the image at each adaptive stage, but for a fixed overall sampling rate, the number of adaptive sampling stages will be reduced. With reduced adaptive stage number, the resolution of sampling rate will be reduced, limiting the potential of adaptive sampling. 
On the other hand, if the sampling scale factor $\alpha$ is too small, the number of adaptive stages is too big and the speed of the adaptive imaging will be slow. 
In order to choose a proper sampling factor, we conduct parameter analysis upon $\alpha$. The quantitative results
are shown in Tab.~\ref{tab:dif_sampling_scale_factor}, with $\alpha$ ranging from 0.5 to 0.8. As is shown, when $\alpha$ increases from 0.5 to 0.7, the overall performance gradually increases. However, when we increase $\alpha$ from $0.7$ to $0.8$, the reconstructed quality starts to decrease. Therefore, according to our experimental results, we choose $\alpha = 0.7$ to obtain the best performance in our experiments. 

\vspace{-2mm}
\section{Discussion}
\label{sec:discussion}
\vspace{-2mm}
In this paper, we propose an adaptive compressive sensing method which could adaptively adjust the sampling region among stages and realize state-of-the art performance. Since the adaptive sampling module requires the reconstruction results of the previous stage as one of the inputs, the inferencing of the reconstruction neural network will be the bottleneck in the real-time CS imaging. However, we need to note that the main contribution of the proposed method in this paper is to develop an highly-efficient adaptive CS imaging principle, based upon the RIP condition theory, and we believe that through adopting techniques such as network compression, quantization, or pruning~\cite{liang2021pruning}, the inferencing speed could be greatly accelerated and the proposed method could be readily applied for real-time adaptive CS imaging. We will leave the network acceleration of the proposed method as further work.

\vspace{-2mm}
\section{Conclusion}
\label{sec:conclusion}
\vspace{-2mm}
In conclusion, we propose an adaptive and cascaded CS method, which can adaptively sample natural images with scene-dependent information, based on the RIP theory. 
Cascaded feature fusion between stages are proposed for CS reconstruction, which makes full use of the information among stages and elegantly helps to further improve the reconstruction efficiency.
Quantitative and qualitative experiments are conducted and demonstrate the state-of-the-art CS effectiveness of the proposed method.

\clearpage

\bibliographystyle{plainnat}
\bibliographystyle{ieee_fullname}
\bibliography{adacs}

\end{document}